\pdfoutput=1

\documentclass[10pt,twocolumn,letterpaper]{article}

\usepackage{cvpr}
\usepackage{times}
\usepackage{graphicx,epstopdf}
\usepackage{amsmath,eucal,amssymb}
\usepackage{amssymb,bibspacing,url,xspace}

\usepackage[bf,small]{caption}

\usepackage{mdwtab}
\usepackage{multirow}
\usepackage{rotating}

\usepackage[scriptsize]{subfigure}

\usepackage[nospace,compress]{cite}
\usepackage{algorithm2e}
\let\savedalgorithm\algorithm
\let\savedendalgorithm\endalgorithm

\usepackage{algorithm}
\newcommand{\fnorm}[2][2]{\ensuremath{ \left\| #2 \right\|^2_{ \mathrm{#1} } } }
\newcommand\norm[1]{\left\lVert#1\right\rVert}

\newcommand{\st}{{{\rm s.t.}\!:}\xspace}

\newcommand{\argmax}{\operatornamewithlimits{argmax}}

\newcommand{\trace}{\operatornamewithlimits{Tr}}
\newcommand{\half}{{\frac{1}{2}}}

\newcommand{\crf}{{CRF}\xspace}
\newcommand{\mrf}{{MRF}\xspace}
\newcommand{\cnn}{{CNN}\xspace}

\def\Real{\mathbb{R}}

\def\regress{{z}}
\def\bregress{{\bf z}}
\def\pws{{R}}
\def\bpws{{\bf R}}

\def\x{{\bf x}}
\def\y{{\bf y}}

\def\I{{\bf I}}
\def\S{{\bf S}}
\def\D{{\bf D}}
\def\A{{\bf A}}

\def\J{{\bf J}}
\def\bbeta{{\boldsymbol \beta}}
\def\btheta{{\boldsymbol \theta}}

\def\T{{\mathrm{T}}}
\def\Pr{{\mathrm{Pr}}}

\def\zeros{{\boldsymbol  0}}

\def\T{{\!\top}}

\cvprfinalcopy %

\begin{document}

\title{Deep Convolutional Neural Fields for Depth Estimation from a Single Image\thanks{
    This work is in part supported by ARC Grants FT120100969, LP120200485, LP130100156.
    Correspondence should be addressed to C. Shen (email: chhshen@gmail.com).
  }
}

\author{
Fayao Liu$^1$,
Chunhua Shen$^{1,2}$,
Guosheng Lin$^{1,2}$
\\
$^1$The University of Adelaide, Australia
\\
$^2$Australian Centre for Robotic Vision
}

\maketitle

\begin{abstract} 
We consider the problem of depth estimation from a single monocular image in this work.
It is a challenging task as no reliable depth cues are available, \eg, stereo correspondences, motions \etc.
Previous efforts have been focusing on exploiting geometric priors or additional sources of information, with all using hand-crafted features.
Recently, there is mounting evidence that features from deep convolutional neural networks (\cnn)  are setting new records for various vision applications.
On the other hand, considering the continuous characteristic of the depth values,  depth  estimations can be naturally formulated into a continuous conditional random field (\crf) learning problem.
Therefore, we in this paper present a deep convolutional neural field model for estimating depths from a single image, aiming to jointly explore the capacity of deep \cnn and continuous \crf.
Specifically, we propose a deep structured learning scheme which learns the unary and pairwise potentials of continuous \crf in a unified deep \cnn framework.

The proposed method can be used for depth estimations of general scenes with no geometric priors nor any extra information injected.
In our case, the integral of the partition function 
can be analytically calculated, thus we can exactly solve the log-likelihood optimization.  
Moreover, solving the MAP problem for predicting depths of a new image is highly efficient as closed-form solutions exist.
We experimentally demonstrate that the proposed method outperforms state-of-the-art depth estimation methods on both indoor and outdoor scene datasets.
\end{abstract}

\section{Introduction}
Estimating depths from a single monocular image depicting general scenes is a fundamental problem in computer vision, which has found wide applications in scene understanding, 3D modelling, robotics, \etc.  
It is a notoriously ill-posed problem, as one captured image may correspond to numerous real world scenes \cite{dcnn_nips14}.
Whereas for humans, inferring the underlying 3D structure from a single image is of little difficulties, it remains a challenging task for computer vision algorithms as no reliable cues can be exploited, such as temporal information, stereo correspondences, \etc. 
Previous works mainly focus on enforcing geometric assumptions, \eg, box models, to infer the spatial layout of a room \cite{Hedau_eccv10,Lee_nips10} or outdoor scenes \cite{Gupta_eccv10}.  
These models come with innate restrictions, which are limitations  to model only particular scene structures and therefore not applicable for general scene depth estimations. 
Later on, non-parametric methods \cite{depthTransfer_pami14} are explored, which consists of candidate images retrieval, scene alignment  and then depth infer using optimizations with smoothness constraints.
This is based on the assumption that scenes with semantic similar appearances should have similar depth distributions when densely aligned.
However, this method is prone to propagate errors through the different decoupled stages and relies heavily on building a reasonable sized image database to perform the candidates retrieval.
In recent years, efforts have been made towards incorporating additional sources of information, \eg, user annotations \cite{Russell_cvpr09}, semantic labellings \cite{Liu_cvpr12, Ladicky_cvpr14}. 
In the recent work of \cite{Ladicky_cvpr14}, Ladicky \etal have shown that jointly performing depth estimation and semantic labelling can benefit each other.
However, they do need to hand-annotate the semantic labels of the images beforehand as such ground-truth information are generally not available.
Nevertheless, all these methods use hand-crafted features.

Different from the previous efforts, we propose to  
formulate the depth estimation as a deep continuous \crf learning problem, without relying on any geometric priors nor any extra information. 
Conditional Random Fields (\crf) \cite{Lafferty_icml01} are popular graphical models used for structured prediction. While extensively studied in classification (discrete) domains, \crf has been less explored for regression (continuous) problems.  
One of the pioneer work on continuous \crf can be attributed to \cite{ccrf_nips08}, in which it was proposed for global ranking in document retrieval.
Under certain constraints, they can directly solve the maximum likelihood optimization as the partition function  can be analytically calculated.
Since then, continuous \crf has been applied for solving various structured regression problems, \eg, remote sensing \cite{ccrf_ecai10,ccrf_aaai13}, image denoising \cite{ccrf_aaai13}.
Motivated by all these successes, we here propose to use it for depth estimation,
given the continuous nature of the depth values, and learn the potential functions in a deep convolutional neural network (\cnn).

Recent years have witnessed the prosperity of the deep convolutional neural network (\cnn). 
CNN features have been setting new records for a wide variety of vision applications \cite{astound_cvprW14}.
Despite all the successes in classification problems, 
deep \cnn has been less explored for structured learning problems, \ie, joint training of a deep \cnn and a graphical model, which is a relatively new and not well addressed problem.
To our knowledge, no such model has been successfully used for depth estimations. 
We here bridge this gap by jointly exploring \cnn and continuous \crf.

To sum up, we highlight the main contributions of this work as follows:
\begin{itemize}
\vspace{-.12cm} \item
We propose a deep convolutional neural field model for depth estimations by exploring \cnn and continuous \crf.  
Given the continuous nature of the depth values, the partition function in the probability density function can be analytically calculated, therefore we can directly solve the log-likelihood optimization without any approximations.  The gradients can be exactly calculated in the back propagation training.
Moreover, solving the MAP problem for predicting the depth of a new image is highly efficient since closed form solutions exist.
\vspace{-.12cm} \item
We jointly learn the unary and pairwise potentials of the \crf in a unified deep \cnn framework, which is trained using back propagation.
\vspace{-.12cm} \item
We demonstrate that the proposed method outperforms state-of-the-art results of depth estimation on both indoor and outdoor scene datasets.
\end{itemize}

\section{Related work}
Prior works \cite{Saxena_nips05,make3d_pami09,Liu_cvpr12} 
typically formulate the depth estimation as a Markov Random Field (\mrf) learning problem.
As exact \mrf learning and inference are intractable in general, 
most of these approaches employ approximation methods, \eg, multi-conditional learning (MCL), particle belief propagation (PBP). 
Predicting the depths of a new image is inefficient, taking around 4-5s in \cite{make3d_pami09} and even longer (30s) in \cite{Liu_cvpr12}.
To make things worse, these methods suffer from lacking of flexibility in that \cite{Saxena_nips05,make3d_pami09}  rely on horizontal alignment of images and %
\cite{Liu_cvpr12} requires the semantic labellings of the training data available beforehand. 
More recently, Liu \etal \cite{Miaomiao_cvpr14} propose a discrete-continuous \crf model to take into consideration the relations between adjacent superpixels, \eg, occlusions. They also need to use approximation methods for learning and MAP inference.
Besides, their method relies on image retrievals to obtain a reasonable initialization.  
By contrast, we here present a deep continuous \crf model in which we can directly solve the log-likelihood optimization without any approximations as the partition function can be analytically calculated.
Predicting the depth of a new image is highly efficient since closed form solution exists.
Moreover, our model do not inject any geometric priors nor any extra information.

On the other hand, previous methods \cite{make3d_pami09,Liu_cvpr12,depthTransfer_pami14,Miaomiao_cvpr14,Ladicky_cvpr14} all use hand-crafted features in their work, \eg, texton, GIST, SIFT, PHOG, object bank, \etc.  
In contrast, we learn deep \cnn for constructing unary and pairwise potentials of \crf.
By jointly exploring the capacity  of \cnn and continuous \crf, our method outperforms state-of-the-art methods on both indoor and outdoor scene depth estimations. 
Perhaps the most related work 
is the recent work of \cite{dcnn_nips14}, which is concurrent to our work here. They train two CNNs for depth map prediction from a single image. 
However, our method bears substantial differences from theirs.
They use the \cnn as a black-box by directly regressing the depth map from an input image through convolutions. 
In contrast, we use \crf to explicitly model the relations of neighbouring superpixels, and learn the potentials in a unified \cnn framework. 
One potential drawback of the method in \cite{dcnn_nips14} is that it tends to learn depths with location preferences, which is prone to fit into specific layouts. 
This partly explains why they have to collect a large number of labelled data to cover all possible layouts for training the networks 
(they collect extra training images using depth sensors)
 , which is in the millions as reported in \cite{dcnn_nips14}.
Instead,  our method enjoys translation invariance as 
we do not encode superpixel coordinates into the unary potentials, 
and can train on standard dataset to get competetive performance without using additional training data. 
Furthermore, the predicted depth map of \cite{dcnn_nips14} is 1/4-resolution of the original input image with some border areas lost, 
while our method does not have this limitation.

In the most recent work of \cite{Lecun_nips14}, Tompson \etal present a hybrid architecture for jointly training a deep \cnn and an \mrf for human pose estimation. 
They first train a unary term and a spatial model separately, then jointly learn them as a fine tuning step.  
During fine tuning of the whole model, they simply remove the partition function in the likelihood to have a loose approximation.
In contrast, our model performs continuous variables prediction. 
We can directly solve the log-likelihood optimization without using approximations as the partition function is integrable and can be analytically calculated. 
Moreover, during prediction, we have closed-form solution for the MAP inference.

\begin{figure*}[!t] \center
	\includegraphics[width=0.88\textwidth]{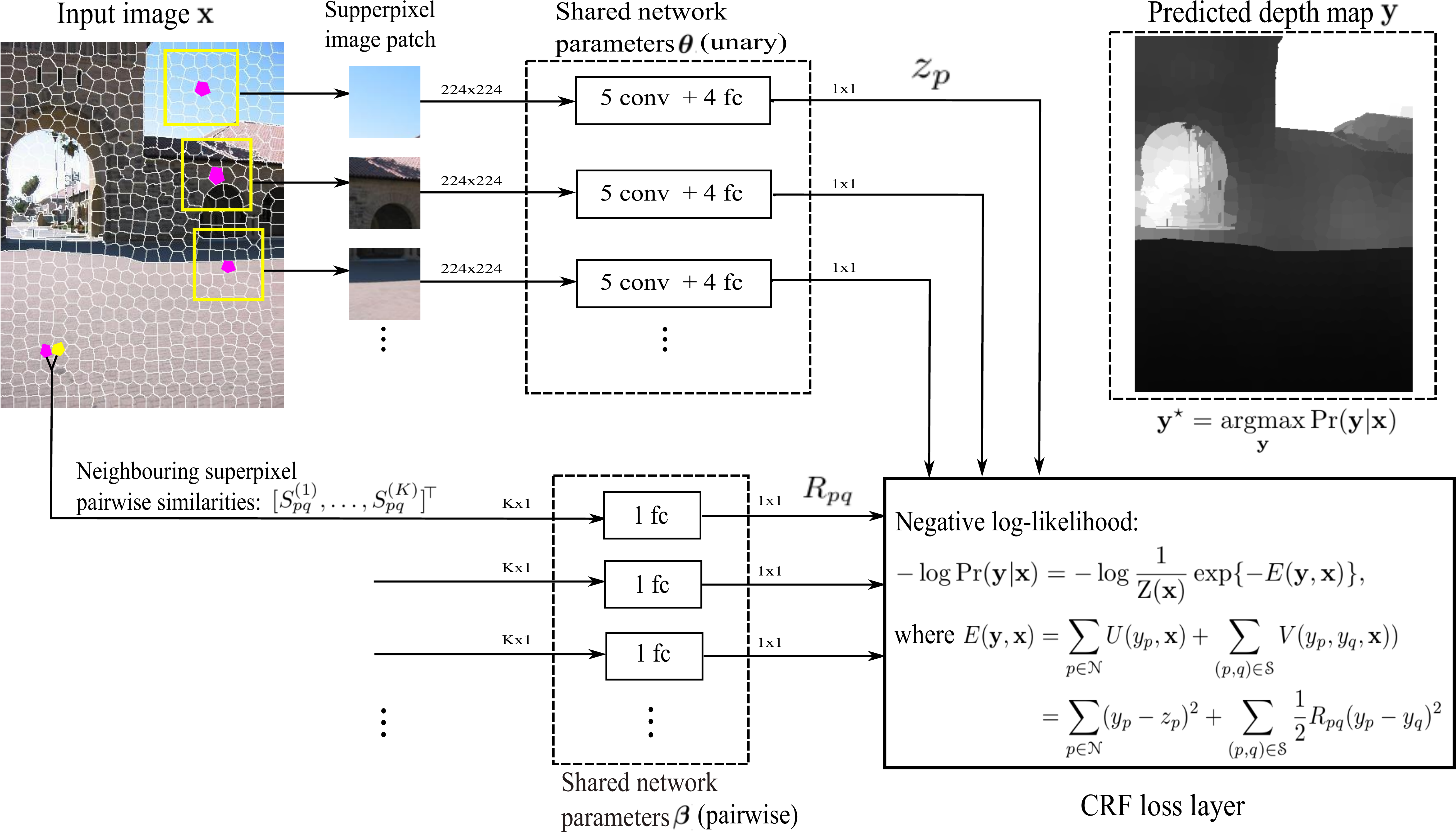}
\caption{An illustration of our deep convolutional neural field model for depth estimation. 
The input image is first over-segmented into superpixels. 
In the unary part, for a superpixel $p$, we crop the image patch centred around its centroid, 
then resize and feed it to a \cnn which is composed of 5 convolutional and 4 fully-connected layers (details refer to Fig. \ref{fig:cnn_unary}).
In the pairwise part,
for a pair of neighbouring superpixels $(p, q)$, we consider $K$ types of similarities, and feed them into a fully-connected layer. 
The outputs of unary part and the pairwise part are then fed to the \crf structured loss layer, which minimizes the negative log-likelihood.     
Predicting the depths of a new image $\x$ is to maximize the conditional probability $\Pr(\y|\x)$, which has closed-form solutions (see Sec. \ref{sec:learning} for details).
}  \label{fig:cnn_ccrf}
\end{figure*}

\section{Deep convolutional neural fields}

We present the details of our deep convolutional neural field model for depth estimation in this section. Unless otherwise stated, we use boldfaced uppercase and lowercase letters to denote matrices and column vectors respectively.
\subsection{Overview}
The goal here is to infer the depth of each pixel in a single image depicting general scenes. 
Following the work of \cite{make3d_pami09,Liu_cvpr12,Miaomiao_cvpr14}, 
we make the common assumption that an image is composed of small homogeneous regions (superpixels)   
and consider the graphical model composed of nodes defined on superpixels.
Kindly note that our framework is flexible that can work on pixels or superpixels.
Each superpixel is portrayed by the depth of its centroid.
Let $\x$ be an image and $\y=[y_1,\ldots, y_n]^{\T} \in \Real^n$ be 
a vector of continuous depth values corresponding to
 all $n$ superpixels in $\x$. 
 Similar to conventional CRF,
 we model the conditional probability distribution of the data with the following density function:
\begin{equation}\label{eq:prob}
\begin{aligned}
\Pr(\y|\x) = \frac{1}{\mathrm{Z(\x)}} \exp (- E(\y, \x)),
\end{aligned}
\end{equation}
where $E$ is the energy function; $\mathrm{Z}$ is the partition function  defined as:
\begin{equation}\label{eq:partition}
\mathrm{Z}(\x) = \int_{\y} \exp \{ -E(\y, \x) \}\mathrm{d}\y.
\end{equation}
Here, because $\y$ is continuous, the integral in Eq. \eqref{eq:prob} can be analytically calculated under certain circumstances, which we will show in Sec. \ref{sec:learning}.
This is different from the discrete case, in which approximation methods need to be applied.  
To predict the depths of a new image, we solve the maximum a posteriori (MAP) inference problem: 
\begin{align} \label{eq:inference}
\y^{\star}&=\argmax_{\y} \Pr(\y|\x). 
\end{align}

\begin{figure*}[!t] \center
	\includegraphics[width=0.85\textwidth]{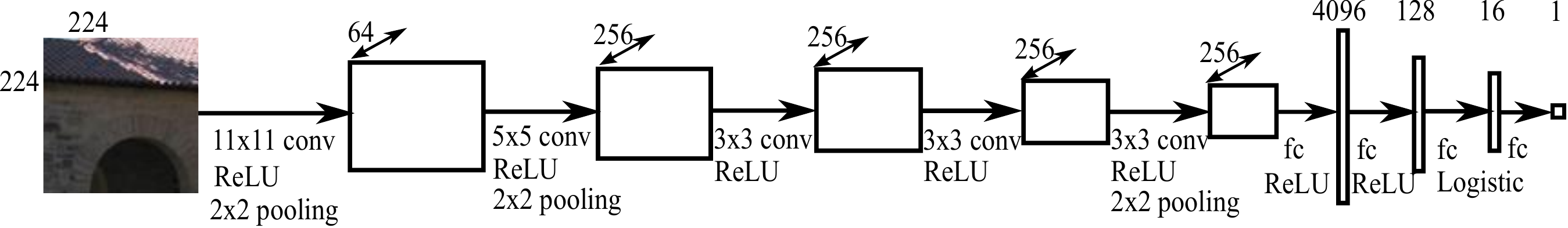}
\caption{Detailed network architecture of the unary part in Fig. \ref{fig:cnn_ccrf}. }  \label{fig:cnn_unary}
\end{figure*}

We formulate the energy function as a typical combination of unary potentials $U$ and pairwise potentials $V$ over the nodes (superpixels) $\cal N$ and edges $\cal S$ of the image $\x$:
\begin{equation}\label{eq:energy}
E(\y, \x) = \sum_{p \in {\cal N} } U(y_{p}, \x) 
	 + \sum_{(p,q) \in {\cal S}} V(y_{p}, y_{q}, \x).
\end{equation}
The unary term $U$ aims to regress the depth value from a single superpixel. 
The pairwise term $V$ encourages neighbouring superpixels with similar appearances to take similar depths.
We aim to jointly learn $U$ and $V$ in a unified \cnn framework.

In Fig. \ref{fig:cnn_ccrf}, we show a sketch of our 
deep convolutional neural field model
for depth estimation.   
As we can see, the whole network is composed of a unary part, a pairwise part and a \crf loss layer.
For an input image, which has been over-segmented into $n$ superpixels, 
we consider image patches centred around each superpxiel centroid.
The unary part then takes  all the image patches as input and feed each of them to a \cnn 
and output an $n$-dimentional vector containing regressed depth values of the $n$ superpixels. 
The network for the unary part is composed of $5$ convolutional and $4$ fully-connected layers with details in Fig. \ref{fig:cnn_unary}. Kindly note that the \cnn parameters are shared across all the superpixels. 
The pairwise part %
takes similarity vectors (each with $K$ components)  of all neighbouring superpixel pairs as input and feed each of them to a fully-connected layer (parameters are shared among different pairs), then output a vector containing all the $1$-dimentional similarities for each of the neighbouring superpixel pair.
The \crf loss layer takes as input the outputs from the unary and the pairwise parts to minimize the negative log-likelihood.
Compared to the direct regression method in \cite{dcnn_nips14}, 
our model possesses two potential advantages: 
1) We achieve translation invariance as we construct unary potentials irrespective of the superpixel's coordinate (shown in Sec. \ref{sec:potentials});  
2) We explicitly model the relations of neighbouring superpixels through pairwise potentials.

In the following, we describe the details of potential functions involved in the energy function in Eq. \eqref{eq:energy}.

\subsection{Potential functions}
\label{sec:potentials}

\paragraph{Unary potential}
The unary potential is constructed from the output of a \cnn by considering the least square loss:
\begin{equation}\label{eq:unary}
U(y_{p}, \x;\btheta) = (y_p - \regress_p (\btheta))^2, \;\; \forall p=1,...,n.
\end{equation}
Here $\regress_p$ is the regressed depth of the superpixel $p$ parametrized by the \cnn parameters $\btheta$.

The network architecture for the unary part is depicted in Fig. \ref{fig:cnn_unary}.
Our \cnn model in Fig. \ref{fig:cnn_unary} is mainly based upon the well-known network architecture of Krizhevsky \etal \cite{AlexNet12} with modifications. 
It is composed of $5$ convolutional layers and $4$ fully connected layers. 
The input image is first over-segmented into superpixels, then for each superpixel, we consider the image patch centred around its centroid. Each of the image patches is resized to $ 224 \times 224$ pixels and then fed to the convolutional neural network.
Note that the convolutional and the fully-connected layers are shared across all the image patches of different superpixels. 
Rectified linear units (ReLU) are used as activiation functions for the five convolutional layers and the first two fully connected layers. 
For the third fully-connected layer, we use the logistic function ($f(x)=(1+e^{-x})^{-1}$) as activiation function.
The last fully-connected layer plays the role of model ensemble with no activiation function followed. 
The output is an $1$-dimentional real-valued depth for a single superpixel.

\paragraph{Pairwise potential}
We construct the pairwise potential from $K$ types of similarity observations, each of which enforces smoothness by exploiting consistency information of neighbouring superpixels:
\begin{align} \label{eq:pairwise}
&V(y_{p}, y_{q}, \x; \bbeta) = \half \pws_{pq} (y_p - y_q)^2, \;\forall p,q=1,...,n.
\end{align}
Here $\pws_{pq}$ is the output of the network in the pairwise part (see  Fig. \ref{fig:cnn_ccrf}) from a neighbouring superpixel pair $(p,q)$. 
We use a fully-connected layer here:
\begin{align}  \label{eq:def_R}
\pws_{pq} = \bbeta^{\T} [S_{pq}^{(1)},\ldots,S_{pq}^{(K)}]^{\T} = \sum_{k=1}^K \beta_k S_{pq}^{(k)},
\end{align} 
where $\S^{(k)}$ is the $k$-th similarity matrix whose elements are $S_{pq}^{(k)}$ ($\S^{(k)}$ is symmetric);
$\bbeta=[\beta_1, \ldots, \beta_k]^{\T}$ 
are the network parameters.
From Eq. \eqref{eq:def_R}, we can see that we don't use any activiation function.
However, as our framework is general, more complicated networks can be seamlessly incorporated for the pairwise part.
In Sec .\ref{sec:learning}, we will show that we can derive a general form for calculating the gradients with respect to $\bbeta$ (see Eq. \eqref{eq:derive_beta_final}).  
To guarantee 
$Z(x)$ (Eq. \eqref{eq:partition}) is integrable, we require $\beta_k \geq 0$ \cite{ccrf_nips08}.

We consider $3$ types of pairwise similarities, measured by the color difference, color histogram difference and texture disparity in terms of local binary patterns (LBP) \cite{LBP_icpr94}, which take the conventional form: 
$S_{pq}^{(k)} = e^{-\gamma \norm{s_p^{(k)} - s_q^{(k)}}}, k=1,2,3$,
where $s_p^{(k)}$, $s_q^{(k)}$ are the observation values of the superpixel $p$, $q$ calculated from color, color histogram and LBP; 
$\norm{\cdot}$ denotes the $\ell_2$ norm of a vector and $\gamma$ is a constant.

\subsection{Learning}
\label{sec:learning}
With the unary and the pairwise pontentials defined in Eq. \eqref{eq:unary}, \eqref{eq:pairwise}, we can now write the energy function as:
\begin{equation}\label{eq:feature}
E(\y, \x) = \sum_{p \in {\cal N} } (y_p - \regress_p)^2 
+ \sum_{(p,q) \in {\cal S}} \half \pws_{pq} (y_p - y_q)^2.
\end{equation}
For ease of expression, we introduce the following notation: 
\begin{align} \label{eq:def_Amat}
 \A = \I+ \D - \bpws, 
\end{align}
where $\I$ is the $ n \times n$ identity matrix; $\bpws$ is the matrix composed of $\pws_{pq}$;  $\D$ is a diagonal matrix with $\D_{pp} = \sum_q \pws_{pq}$.
Expanding Eq. \eqref{eq:feature}, we have:
\begin{align} \label{eq:feature_expand}
E(\y, \x) = \y^\T \A \y - 2\bregress^\T \y + \bregress^\T \bregress.
\end{align}
Due to the quadratic terms of $\y$ in the energy function in Eq. \eqref{eq:feature_expand} and the positive definiteness of $\A$, we can analytically calculate the integral in the partition function (Eq. \eqref{eq:partition}) as:
\begin{align} \label{eq:partition_expand}
Z(\x) &= \int_{\y} \exp \{ -E(\y, \x) \}\mathrm{d}\y  \notag \\
 	&=\frac{{(\pi)}^{\frac{n}{2}}}{{|\A|}^{\frac{1}{2}}} \exp \{{ \bregress ^\T \A^{-1} \bregress} - \bregress^\T \bregress \}.
\end{align}
From Eq. \eqref{eq:prob}, \eqref{eq:feature_expand},  \eqref{eq:partition_expand},  we can now write the probability distribution function as (see supplementary for details):
\begin{align} \label{eq:prob_final}
\Pr(\y|\x) =\frac{ |\A|^{\half}} {{\pi^{\frac{n}{2}}}} \exp \Big\{ - \y^\T \A \y + 2\bregress^\T \y -  \bregress ^\T \A^{-1} \bregress \Big\}, 
\end{align}
where $\bregress=[\regress_1, \ldots, \regress_n]^{\T}$; $|\A|$ denotes the determinant of the matrix $\A$, and $\A^{-1}$ the inverse of $\A$.
Then the negative log-likelihood can be written as:
\begin{align} \label{eq:log-likelihood}
-\log\Pr(\y|\x)&= \y^\T \A \y - 2\bregress^\T \y + \bregress ^\T \A^{-1} \bregress  \\ \notag
& - \half\log(|\A|) + \frac{n}{2}\log(\pi). 
\end{align}

During learning, we minimizes the negative conditional log-likelihood of the training data. Adding regularization to $\btheta$, $\bbeta$, we then arrive at the final optimization:
\begin{align} \label{eq:ccrf_final}
\min_{\btheta, \bbeta \geq \zeros} &-\sum_{i=1}^N \log\Pr(\y^{(i)} | \x^{(i)}; \btheta, \bbeta) \\ \notag 
&+ \frac{\lambda_1}{2} \fnorm \btheta + \frac{\lambda_2}{2} \fnorm \bbeta,
\end{align} 
where $\x^{(i)}$, $\y^{(i)}$ denote the $i$-th training image and the corresponding depth map; $N$ is the number of training images; $\lambda_1$ and $\lambda_2$ are weight decay parameters.
We use stochastic gradient descent (SGD) based back propagation to solve the optimization problem in Eq.  \eqref{eq:ccrf_final} for learning all parameters of the whole network.
We project the solutions to the feasible set when the bounded constraints $\beta_k \geq 0$ is violated.
In the following, we calculate the partial derivatives of $-\log\Pr(\y|\x)$ with respect to the network parameters $\theta_l$ (one element of $\btheta$) and $\beta_k$ (one element of $\bbeta$) by using the chain rule (refer to supplementary for details): 
\begin{align}   
\frac{\partial \{ -\log\Pr(\y|\x)\} }{\partial \theta_l} 
& = 2 (\A^{-1} \bregress - \y)^{\T} \frac{\partial \bregress}{\partial \theta_l}, \label{eq:derive_z} \\
\frac{\partial \{ -\log\Pr(\y|\x)\} }{\partial \beta_k} 
& = \y^{\T} \J \y - \bregress^{\T} \A^{-1} \J \A^{-1}\bregress \notag \\
& \; - \half \trace \Big( \A^{-1} \J \Big),  \label{eq:derive_beta_final}
\end{align}
where $\trace(\cdot)$ denotes the trace of a matrix; $\J$ is an $n \times n$ matrix with elements:
\begin{align}  \label{eq:def_J}
J_{pq}& =  - \frac{\partial  \pws_{pq} }{\partial \beta_k} 
+ \delta(p=q) \sum_q \frac{\partial  \pws_{pq} }{\partial \beta_k}, 
\end{align}
where $\delta(\cdot)$ is the indicator function, which equals 1 if $p=q$ is true and 0 otherwise. 
From Eq. \eqref{eq:def_J}, we can see that our framework is general and more complicated networks for the pairwise part can be seamlessly incorporated.
Here, in our case, with the definition of $\pws_{pq}$ in Eq. \eqref{eq:def_R}, we have $\frac{\partial  \pws_{pq} }{\partial \beta_k}=S_{pq}^{(k)}$.

\paragraph{Depth prediction}
Predicting the depths of a new image is to solve the MAP inference in Eq. \eqref{eq:inference}, in which closed form solutions exist here (details refer to supplementary): %
\begin{align} \label{eq:inf_solution}
\y^{\star}&=\argmax_{\y} \Pr(\y|\x)  \\ \notag
&= \argmax_{\y} -\y^\T \A \y + 2\bregress^\T \y  \\  \notag
&= \A^{-1} \bregress.
\end{align} 
If we discard the pairwise terms, namely $\pws_{pq}=0$, then Eq. \eqref{eq:inf_solution} degenerates to $\y^{\star}=\bregress$, which is a conventional regression model (we will report the results of this method as a baseline in the experiment).

\subsection{Implementation details}
We implement the network training based on the efficient CNN toolbox: VLFeat MatConvNet\footnote{VLFeat MatConvNet: http://www.vlfeat.org/matconvnet/} \cite{matconvnet}. 
Training is done on a standard desktop with an NVIDIA GTX 780 GPU with 6GB memory. 
During each SGD iteration, around $\sim 700$ superpixel image patches need to be processed.
The 6GB GPU may not be able to process all the image patches at one time.
We therefore partition the superpixel image patches of one image into two parts and process them successively.
Processing one image takes around $10$s (including forward and backward) with $\sim 700$ superpixels when training the whole network.

During implementation, we initialize the first 6 layers of the unary part in Fig. \ref{fig:cnn_unary} using a \cnn model trained on the ImageNet from \cite{vgg_bmvc14}.
First, we do not back propagate through the previous 6 layers by keeping them fixed and train the rest of the network (we refer this process as pre-train) with the following settings: 
momentum is set to $0.9$, and weight decay parameters $\lambda_1, \lambda_2$ are set to $0.0005$.
During pre-train, the learning rate is initialized at $0.0001$,
and decreased by 40\% every 20 epoches. 
We then run 60 epoches to report the results of pre-train (with learning rate decreased twice).
The pre-train is rather efficient, taking around $1$ hour to train on the Make3D dataset, and inferring the depths of a new image takes less than $0.1$s.
Then we train 
the whole network with the same momentum and weight decay.
We apply dropout with ratio $0.5$ in the first two fully-connected layers of Fig. \ref{fig:cnn_unary}. 
Training the whole network takes around $16.5$ hours on the Make3D dataset, and around $33$ hours on the NYU v2 dataset. Predicting the depth of a new image from scratch takes $\sim 1.1$s.

 \begin{figure*} [t]  \center
\resizebox{.9\linewidth}{!} {
\begin{tabular}{cccccc}
\rotatebox{90}{Test image}
	\includegraphics[width=0.16\textwidth]{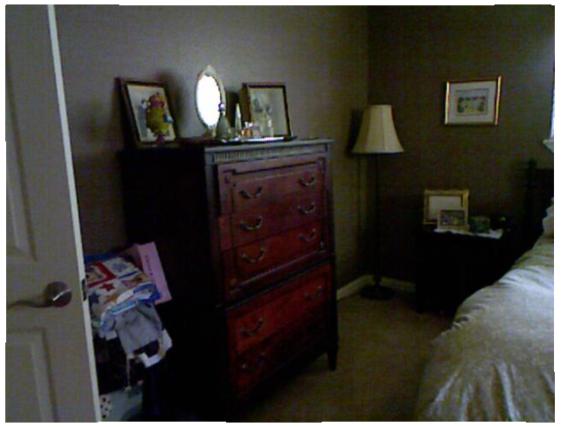}
    \includegraphics[width=0.16\textwidth]{./fig/NYUD2/images/5.jpg}
    \includegraphics[width=0.16\textwidth]{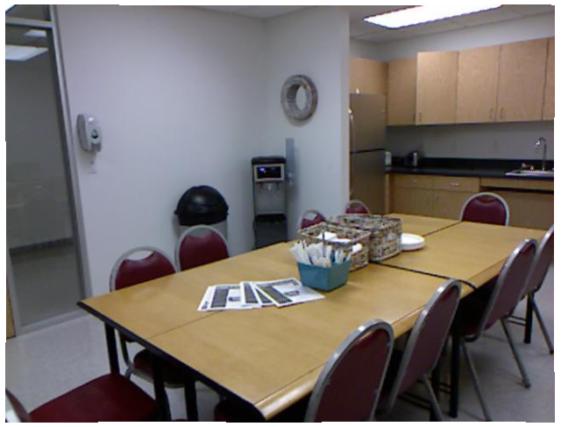}
    \includegraphics[width=0.16\textwidth]{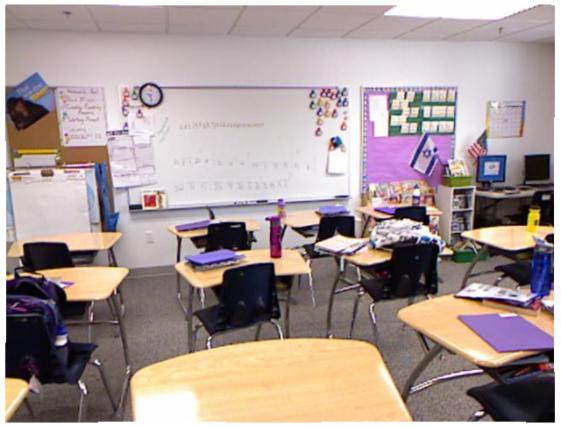}
    \includegraphics[width=0.16\textwidth]{./fig/NYUD2/images/8.jpg}
    \includegraphics[width=0.16\textwidth]{./fig/NYUD2/images/11.jpg} \\
     
\rotatebox{90}{Ground-truth}    
     \includegraphics[width=0.16\textwidth]{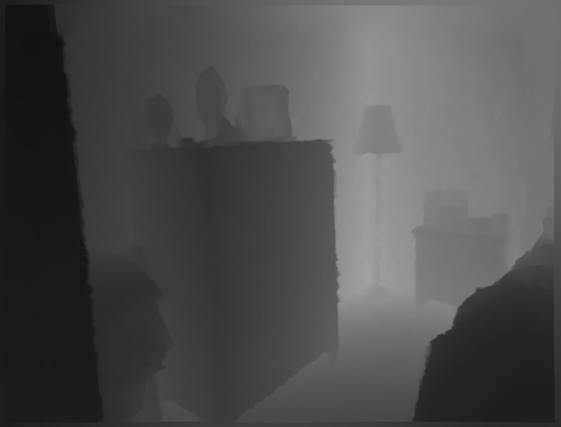}
     \includegraphics[width=0.16\textwidth]{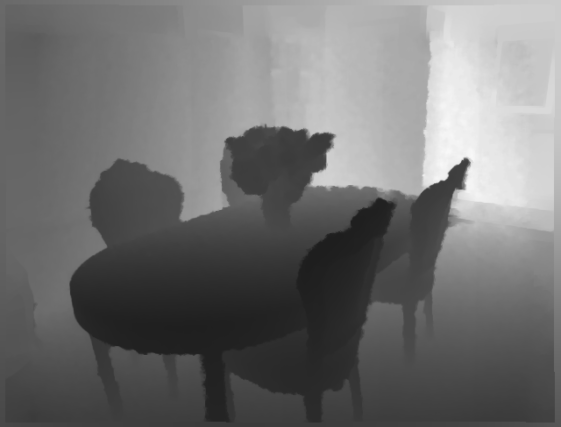}
     \includegraphics[width=0.16\textwidth]{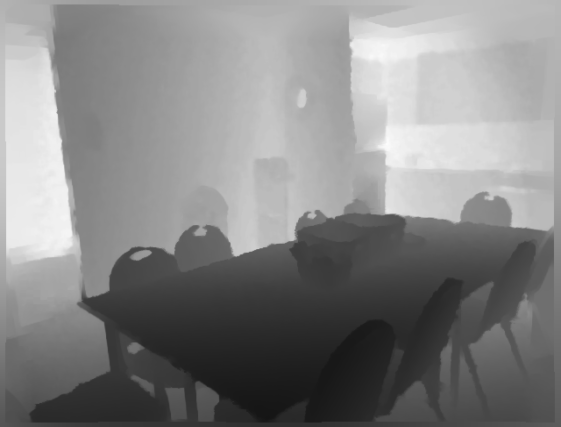}
     \includegraphics[width=0.16\textwidth]{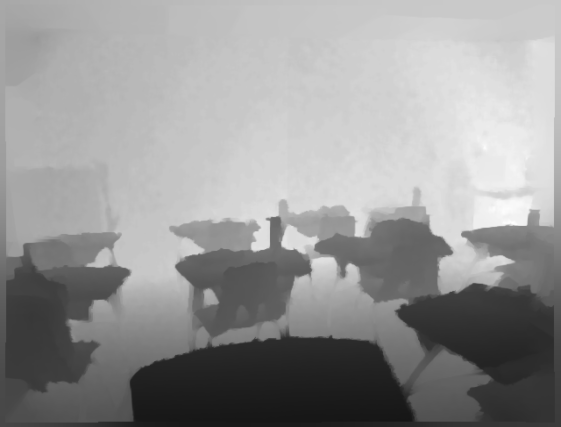}
     \includegraphics[width=0.16\textwidth]{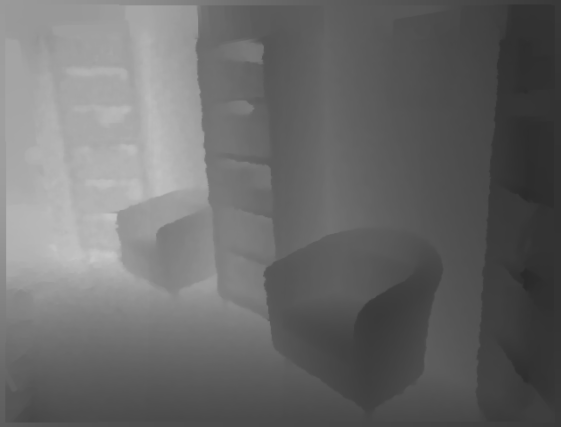}
     \includegraphics[width=0.16\textwidth]{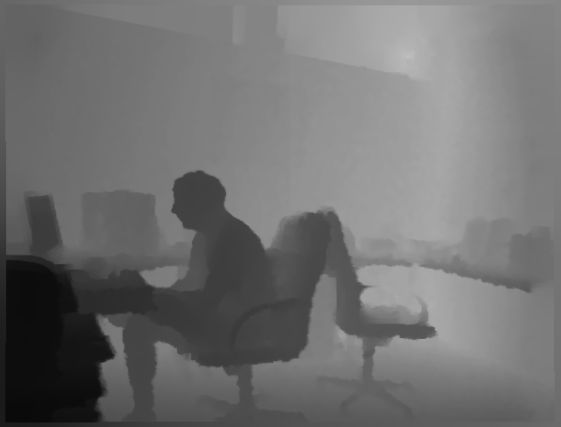} \\
     
\rotatebox{90}{Eigen \etal \cite{dcnn_nips14}}     
     \includegraphics[width=0.16\textwidth]{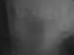} 
     \includegraphics[width=0.16\textwidth]{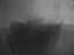} 
     \includegraphics[width=0.16\textwidth]{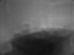}
     \includegraphics[width=0.16\textwidth]{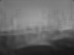} 
     \includegraphics[width=0.16\textwidth]{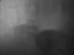} 
     \includegraphics[width=0.16\textwidth]{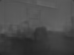}  \\

\rotatebox{90}{Ours (fine-tune)}     
     \includegraphics[width=0.16\textwidth]{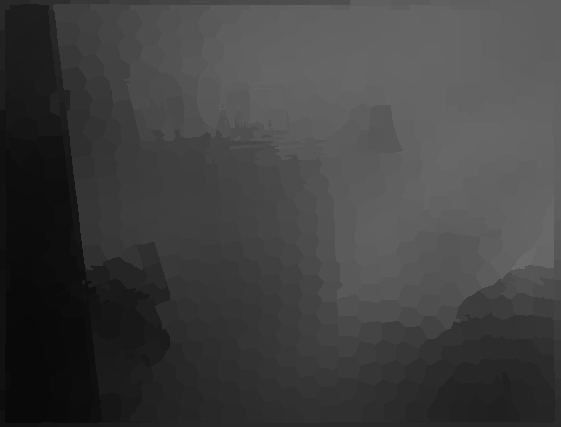}  
     \includegraphics[width=0.16\textwidth]{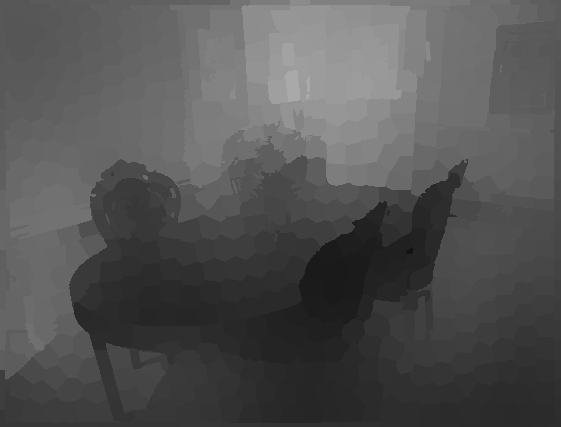}
     \includegraphics[width=0.16\textwidth]{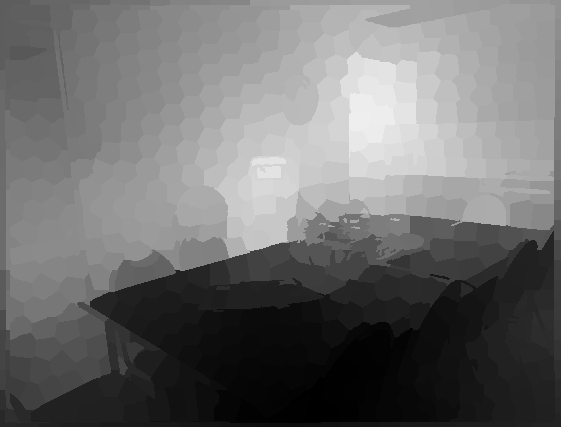}
     \includegraphics[width=0.16\textwidth]{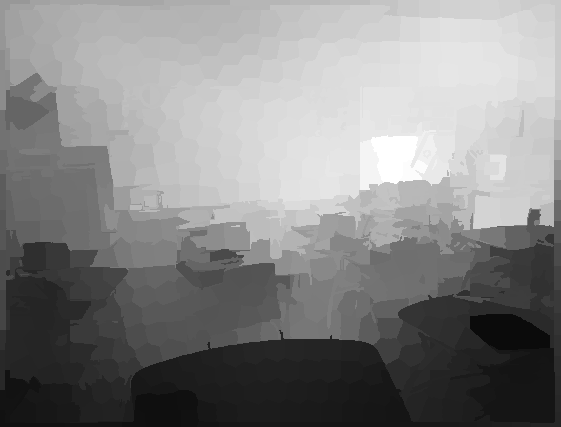}    
     \includegraphics[width=0.16\textwidth]{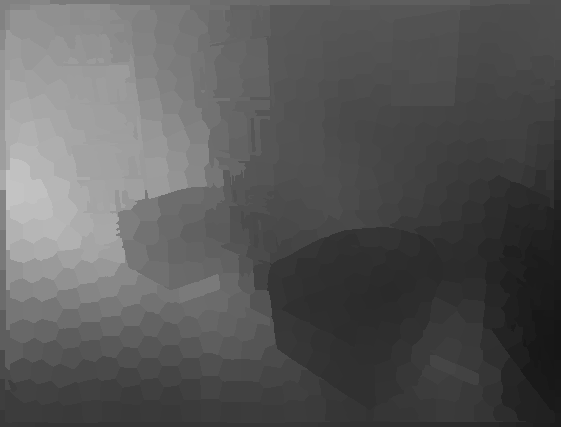}
     \includegraphics[width=0.16\textwidth]{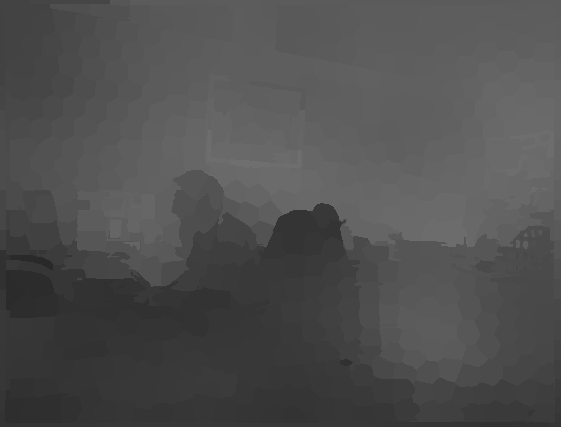}  \\   
\end{tabular}  
}   
\caption{Examples of qualitative comparisons on the NYUD2 dataset (Best viewed on screen).
Our method yields visually better predictions with sharper transitions, aligning to local details.
 }  \label{fig:nyud2}
\end{figure*}

\begin{table*} [t] \center
\resizebox{.65\linewidth}{!} {
\begin{tabular}{ | l |  c  c  c | c  c  c |}
\hline 
\multirow{3}{*}{{{Method}}} &\multicolumn{3}{c|}{Error} &\multicolumn{3}{c|}{Accuracy} \\
&\multicolumn{3}{c|}{(lower is better)} &\multicolumn{3}{c|}{(higher is better)} \\
\cline{2-7}
&rel &log10 &rms &$\delta < 1.25$ &$\delta < 1.25^2$ &$\delta < 1.25^3$  \\
\hline
Make3d \cite{make3d_pami09}		  &0.349  &-  &1.214	&0.447	&0.745	&0.897 \\
DepthTransfer \cite{depthTransfer_pami14}	&0.35 	&0.131	&1.2		&- &- &-	 \\
Discrete-continuous \crf \cite{Miaomiao_cvpr14}	    &0.335	&0.127	&1.06 &- &- &-	 \\
Ladicky \etal \cite{Ladicky_cvpr14}		&- &- &-	&0.542	&0.829	&0.941  \\
Eigen \etal \cite{dcnn_nips14}      &\textbf{0.215}	&-	&0.907	&0.611	&\textbf{0.887}	&\textbf{0.971} \\
\hline
Ours (pre-train) &0.257	 &0.101	  &0.843  	&0.588   	&0.868	&0.961  \\
Ours (fine-tune)    &0.230	 &\textbf{0.095} 	 &\textbf{0.824}  &\textbf{0.614} 	 &0.883	 &\textbf{0.971} \\
\hline
\end{tabular}
}
\caption{Result comparisons on the NYU v2 dataset. 
Our method performs the best in most cases.
Kindly note that the results of Eigen \etal \cite{dcnn_nips14} are obtained by using extra training data (in the millions in total) while ours are obtained using the standard training set.} \label{tab:nyud2}
\end{table*}

\begin{table}
\center
\resizebox{1\linewidth}{!} {
\begin{tabular}{ | l |  c  c  c | c  c  c |}
\hline 
\multirow{3}{*}{{{Method}}} &\multicolumn{3}{c|}{Error} &\multicolumn{3}{c|}{Accuracy} \\
&\multicolumn{3}{c|}{(lower is better)} &\multicolumn{3}{c|}{(higher is better)} \\
\cline{2-7}
&rel &log10 &rms &$\delta < 1.25$ &$\delta < 1.25^2$ &$\delta < 1.25^3$  \\
\hline
SVR  &0.313	&0.128	&1.068	&0.490	&0.787	&0.921 \\
SVR (smooth) &0.290	 &0.116	&0.993	&0.514	&0.821	&0.943\\
Ours (unary only)  &0.295 	 &0.117 	 &0.985   &0.516 	 &0.815 	 &0.938  \\ 
Ours (pre-train) &0.257	 &0.101	  &0.843  	&0.588   	&0.868	&0.961  \\
Ours (fine-tune)  &\textbf{0.230} 	 &\textbf{0.095} 	 &\textbf{0.824}  &\textbf{0.614} 	 &\textbf{0.883} 	 &\textbf{0.971}\\    
\hline
\end{tabular}
}
\caption{Baseline comparisons on the NYU v2 dataset. 
Our method with the whole network training performs the best.
} \label{tab:anal_nyud2}
\end{table}

\begin{table}
\center
\resizebox{0.90\linewidth}{!} {
\begin{tabular}{ | l |  c  c  c | c  c  c |}
\hline 
\multirow{3}{*}{{{Method}}} &\multicolumn{3}{c|}{Error (C1)} &\multicolumn{3}{c|}{Error (C2)} \\
&\multicolumn{3}{c|}{(lower is better)} &\multicolumn{3}{c|}{(lower is better)} \\
\cline{2-7}
&rel &log10 &rms  &rel &log10 &rms  \\
\hline
SVR  &0.433	&0.158	&8.93  &0.429	&0.170	&15.29  \\
SVR (smooth) &0.380	&0.140	&\textbf{8.12}  &0.384	&0.155	&15.10 \\
Ours (unary only) &0.366 	 &0.137 	 &8.63  &0.363 	 &0.148 	 &14.41 \\
Ours (pre-train) &0.331	 &0.127	 &8.82  &0.324	&0.134	&13.29 \\
Ours (fine-tune)   &\textbf{0.314} &\textbf{0.119} &8.60 &\textbf{0.307} &\textbf{0.125} &\textbf{12.89} \\ 
\hline
\end{tabular}
}
\caption{Baseline comparisons on the Make3D dataset. 
Our method with the whole network training performs the best.
} \label{tab:anal_make3d}
\end{table}

\begin{table*} \center
\resizebox{.55\linewidth}{!} {
\begin{tabular}{ | l |  c  c  c | c  c  c |}
\hline 
\multirow{2}{*}{{{Method}}} &\multicolumn{3}{c|}{Error (C1)} &\multicolumn{3}{c|}{Error (C2)} \\
&\multicolumn{3}{c|}{(lower is better)} &\multicolumn{3}{c|}{(lower is better)} \\
\cline{2-7}
&rel &log10 &rms  &rel &log10 &rms  \\
\hline
Make3d \cite{make3d_pami09}		  &-  &-  &- 	&0.370	&0.187  &-  \\
Semantic Labelling \cite{Liu_cvpr12}  &-  &-  &- 	&0.379 &0.148 &- \\
DepthTransfer \cite{depthTransfer_pami14}	&0.355	&0.127	&9.20  &0.361	  &0.148	  &15.10 \\
Discrete-continuous \crf \cite{Miaomiao_cvpr14}	    &0.335	&0.137	&9.49  &0.338	  &0.134	 &\textbf{12.60}  \\
\hline
Ours (pre-train) &0.331	 &0.127	 &8.82  &0.324	&0.134	&13.29 \\
Ours (fine-tune)  &\textbf{0.314} &\textbf{0.119} &\textbf{8.60} &\textbf{0.307} &\textbf{0.125} &12.89 \\
\hline
\end{tabular}
}
\caption{Result comparisons on the Make3D dataset. 
Our method performs the best.
Kindly note that the C2 errors of the Discrete-continuous \crf \cite{Miaomiao_cvpr14} are reported with an ad-hoc post-processing step (train a classifier to label sky pixels and set the corresponding regions to the maximum depth).} \label{tab:make3d}
\end{table*}

\begin{figure*} [t] \center
\resizebox{.80\linewidth}{!} {
\begin{tabular}{cccccc}
\rotatebox{90}{Test image}
	\includegraphics[width=0.12\textwidth]{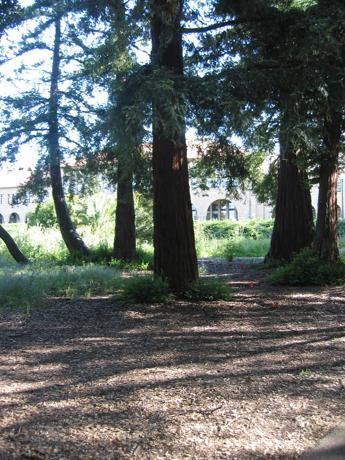}
	\includegraphics[width=0.12\textwidth]{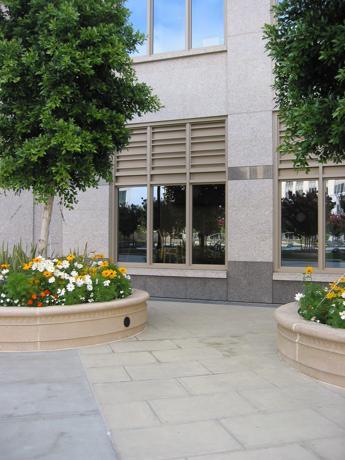}
	\includegraphics[width=0.12\textwidth]{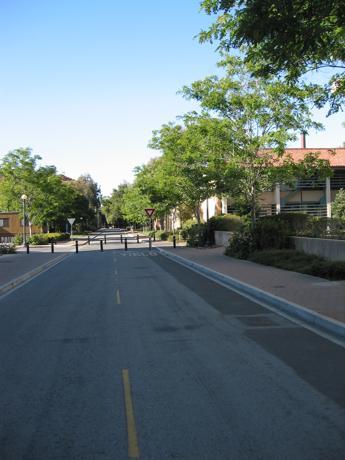}
	\includegraphics[width=0.12\textwidth]{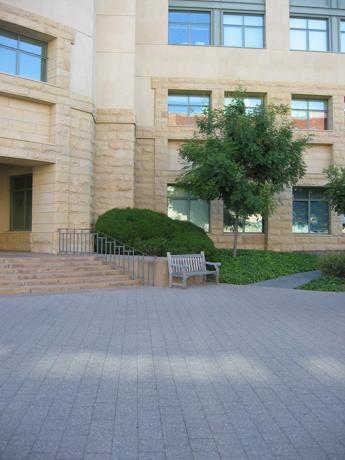}
	\includegraphics[width=0.12\textwidth]{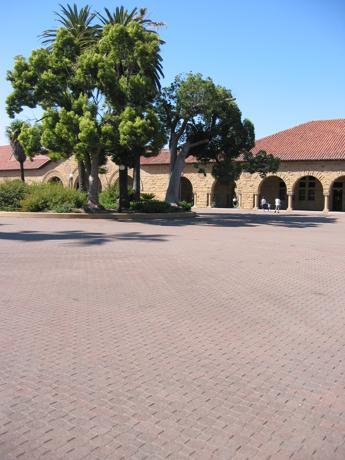}
	\includegraphics[width=0.12\textwidth]{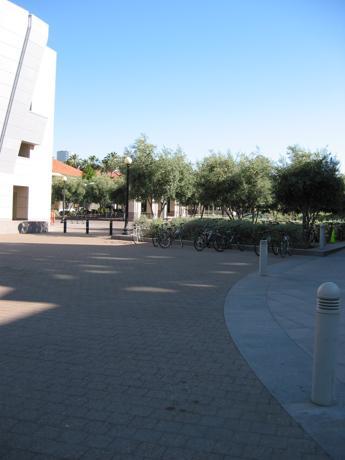} \\

\rotatebox{90}{Ground-truth}
	\includegraphics[width=0.12\textwidth]{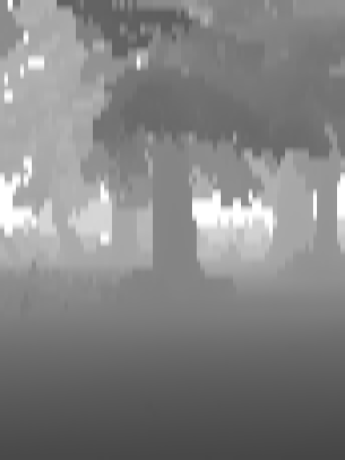}
	\includegraphics[width=0.12\textwidth]{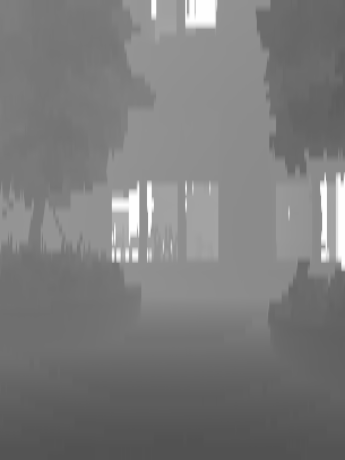}
	\includegraphics[width=0.12\textwidth]{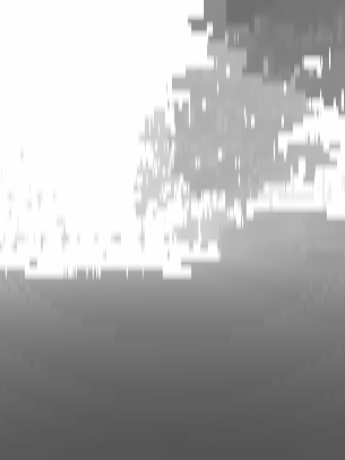}
	\includegraphics[width=0.12\textwidth]{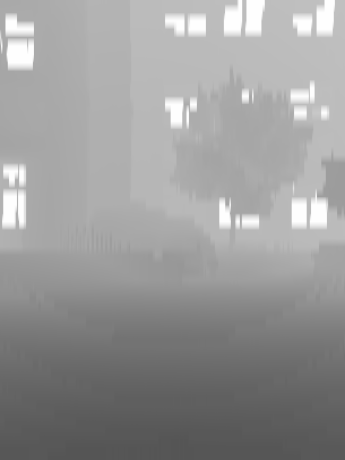}
	\includegraphics[width=0.12\textwidth]{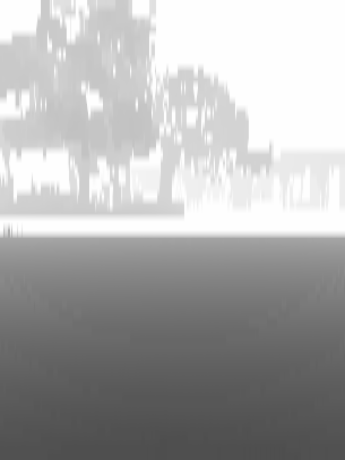}
	\includegraphics[width=0.12\textwidth]{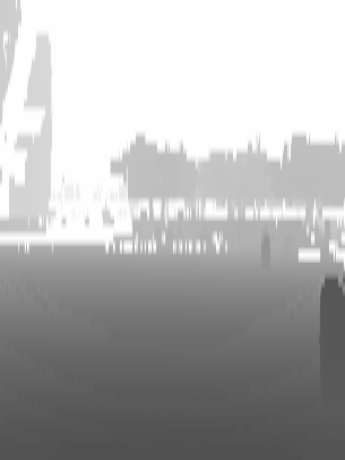} \\

\rotatebox{90}{Ours (unary only)}
	\includegraphics[width=0.12\textwidth]{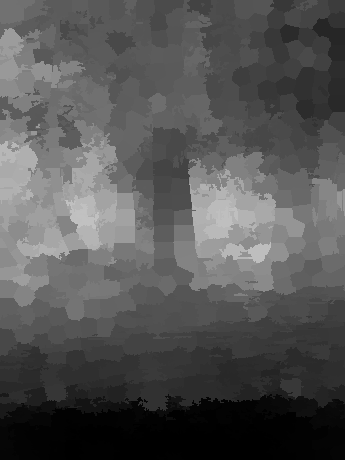}
	\includegraphics[width=0.12\textwidth]{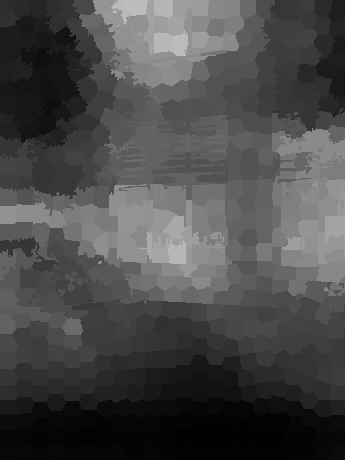}
	\includegraphics[width=0.12\textwidth]{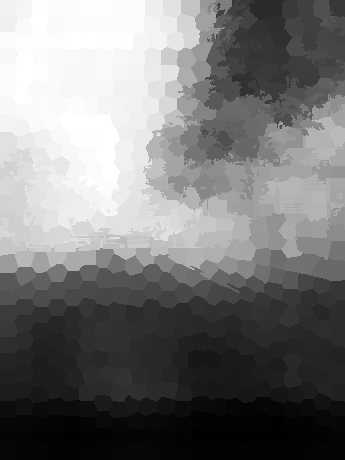}
	\includegraphics[width=0.12\textwidth]{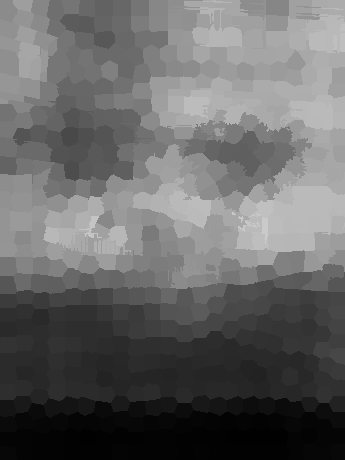}
	\includegraphics[width=0.12\textwidth]{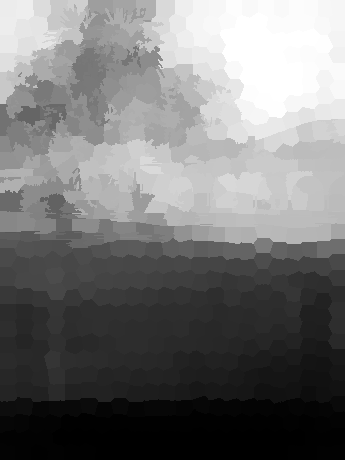}
	\includegraphics[width=0.12\textwidth]{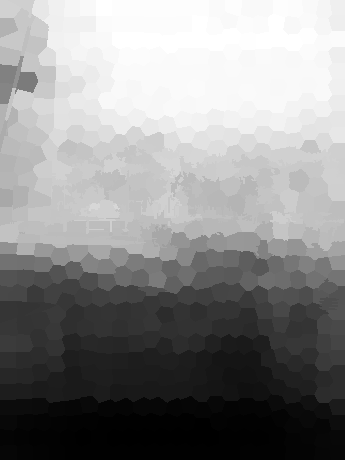} \\

\rotatebox{90}{Ours (fine-tune)}
	\includegraphics[width=0.12\textwidth]{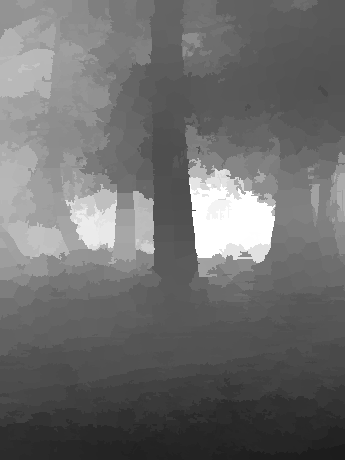}
	\includegraphics[width=0.12\textwidth]{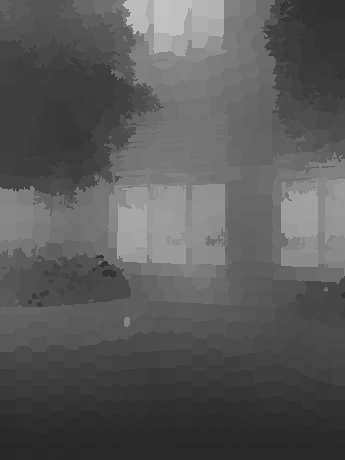}
	\includegraphics[width=0.12\textwidth]{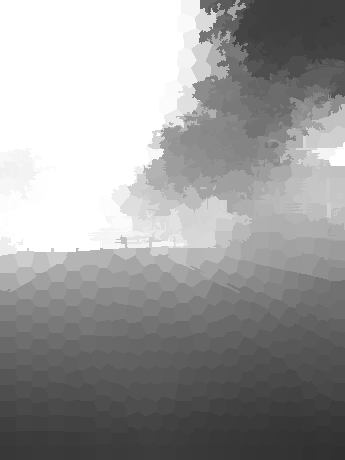}
	\includegraphics[width=0.12\textwidth]{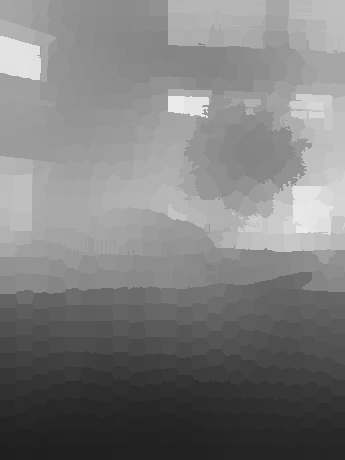}
	\includegraphics[width=0.12\textwidth]{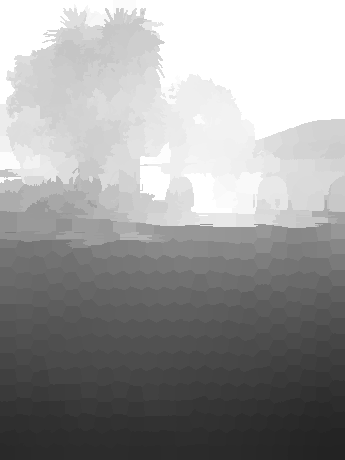}
	\includegraphics[width=0.12\textwidth]{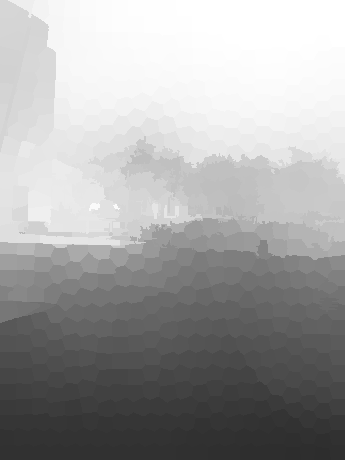} \\	
 \end{tabular}      
 } 
\caption{Examples of depth predictions on the Make3D dataset (Best viewed on screen).
The unary only model gives rather coarse predictions, with blurry boundaries and segments. In contrast, our full model with pairwise smoothness yields much better predictions.
}  \label{fig:nmake3d}
\end{figure*}

\section{Experiments}

We evaluate on two popular datasets which are available online: the NYU v2 Kinect dataset \cite{nyud2_eccv12} and the Make3D range image dataset \cite{make3d_pami09}. 
Several measures commonly used in prior works are applied here for quantitative evaluations:
\begin{itemize}
\vspace{-.12cm} \item
average relative error (rel): 
$\frac{1}{T} \sum_p \frac{|d_p^{gt} - d_p|}{d_p^{gt}} $;
\vspace{-.12cm} \item
root mean squared error (rms): 
$\sqrt{\frac{1}{T} \sum_p (d_p^{gt} - d_p)^2}$;
\vspace{-.12cm} \item
average $\log_{10}$ error (log10): \\
$\frac{1}{T} \sum_p | \log_{10}d_p^{gt} - \log_{10}d_p|$;
\vspace{-.12cm} \item
accuracy with threshold $thr$: \\
percentage ($\%$) of $d_p \; \; \st \max (\frac{d_p^{gt}}{d_p}, \frac{d_p}{d_p^{gt}}) = \delta < thr$;
\end{itemize}
where $d_p^{gt}$ and $d_p$ are the ground-truth and predicted depths respectively at pixel indexed by $p$, and $T$ is the total number of pixels in all the evaluated images. 

We use SLIC \cite{slic_pami12} to segment the images into a set of non-overlapping superpixels. For each superpixel, we consider the image within a rectangular box centred on the centroid of the superpixel, which contains a large portion of its background surroundings. More specifically, we use a box size of 168$\times$168 pixels for the NYU v2 and $120 \times 120$ pixels for the Make3D dataset.
Following \cite{make3d_pami09,Liu_cvpr12,dcnn_nips14}, we transform the depths into log-scale before training.
As for baseline comparisons, we consider the following settings:
\begin{itemize}
\vspace{-.12cm} \item
SVR: We train a support vector regressor using the \cnn representations from the first 6 layers of Fig. \ref{fig:cnn_unary}; 
\vspace{-.12cm} \item
SVR (smooth): We add a smoothness term to the trained SVR during prediction by solving the inference problem in Eq. \eqref{eq:inf_solution}. As tuning multiple pairwise parameters is not straightforward, we only use color difference as the pairwise potential and choose the parameter $\beta$ by hand-tuning on a validation set;
\vspace{-.12cm} \item
Unary only: We replace the \crf loss layer in Fig. \ref{fig:cnn_ccrf} with a least-square regression layer (by setting the pairwise outputs $\pws_{pq}=0$, $p,q=1,...,n$), which degenerates to a deep regression model trained by SGD.
\end{itemize}

\subsection{NYU v2: Indoor scene reconstruction}
The NYU v2 dataset consists of $1449$ RGBD images of indoor scenes, among which $795$ are used for training and 654 for test (we use the standard training/test split provided with the dataset).
Following \cite{Miaomiao_cvpr14}, we resize the images to $427 \times 561$ pixels 
before training.%

For a detailed analysis of our model, we first compare with the three baseline methods and report the results in Table \ref{tab:anal_nyud2}. 
From the table, several conclusions can be made: 1) When trained with only unary term, deeper network is beneficial for better performance, which is demonstrated by the fact that our unary only model outperforms the SVR model;
2) Adding smoothness term to the SVR or our unary only model helps improve the prediction accuracy;
3) Our method achieves the best performance by jointly learning the unary and the pairwise parameters in a unified deep \cnn framework. Moreover, fine-tuning the whole network yields further performance gain.
These well demonstrate the efficacy of our model.

In Table \ref{tab:nyud2}, we compare our model with several popular state-of-the-art methods.  
As can be observed, our method outperforms classic methods like Make3d \cite{make3d_pami09}, DepthTransfer \cite{depthTransfer_pami14} with large margins. 
Most notably, our results are significantly better than that of \cite{Ladicky_cvpr14}, which jointly exploits depth estimation and semantic labelling. 
Comparing to the recent work of Eigen \etal \cite{dcnn_nips14}, our method generally performs on par.
Our method obtains significantly better result in terms of root mean square (rms) error.
Kindly note that, to overcome overfit, they \cite{dcnn_nips14} have to collect millions of additional labelled images to train their model. One possible reason is that their method captures the absolute pixel location information and they probably need a very large training set to cover all possible pixel layouts.
In contrast, we only use the standard training sets ($795$) without any extra data,  
yet we achieve comparable or even better performance.
Fig. \ref{fig:nyud2} illustrates some qualitative evaluations of our method compared against Eigen \etal \cite{dcnn_nips14} (We download the predictions of \cite{dcnn_nips14} from the authors' website.). 
Compared to the predictions of \cite{dcnn_nips14},
our method yields more visually pleasant predictions with sharper transitions, aligning to local details.

\subsection{Make3D: Outdoor scene reconstruction}
The Make3D dataset contains 534 images depicting outdoor scenes.
As pointed out in \cite{make3d_pami09,Miaomiao_cvpr14}, this dataset is with limitations: the maximum value of depths is 81m with far-away objects are all mapped to the one distance of $81$ meters. 
As a remedy, two criteria are used in \cite{Miaomiao_cvpr14} to report the prediction errors: ($C_1$) Errors are calculated only in the regions with the ground-truth depth less than $70$ meters; ($C_2$) Errors are calculated over the entire image.
We follow this protocol to report the evaluation results. 

Likewise, we first present the baseline comparisons in Table \ref{tab:anal_make3d}, from which similar conclusions can be drawn as in the NYU v2 dataset. 
We then show the detailed results compared with several state-of-the-art methods in Table \ref{tab:make3d}.  
As can be observed, our model with 
the whole network training
ranks the first in overall performance,
outperforming the compared methods by large margins.
Kindly note that the C2 errors of \cite{Miaomiao_cvpr14} are reported with an ad-hoc post-processing step, which trains a classifier to label sky pixels and set the corresponding regions to the maximum depth.
In contrast, we do not employ any of those heuristics to refine our results, yet we achieve better results in terms of relative error. %
Some examples of qualitative  evaluations are shown in Fig. \ref{fig:nmake3d}.  
It is shown that our unary only model gives rather coarse predictions with blurry boundaries. 
By adding smoothness term, our model yields much better visualizations, which are close to ground-truth.

\section{Conclusion}
We have presented a deep convolutional neural field model for depth estimation from a single image.
The proposed method combines the strength of deep \cnn and continuous \crf
in a unified \cnn framework.
We show that the log-likelihood optimization in our method can be directly solved using back propagation without any approximations required.
Predicting the depths of a new image by solving the MAP inference can be efficiently performed as closed-form solutions exist.
Given the general learning framework of our method, it can also be applied for other vision applications, \eg, image denoising.
Experimental results demonstrate that the proposed method outperforms state-of-the-art methods on both indoor and outdoor scene datasets.

\appendix

\onecolumn

\numberwithin{equation}{section}

\section{Deep Convolutional Neural Fields}

In this Appendix, we show some technical details about the proposed deep convolutional field model.

Let $\x$ be an image and $\y=[y_1,\ldots, y_n]^{\T} \in \Real^n$ be
a vector of continuous depth values corresponding to
 all $n$ superpixels in $\x$.
 Similar to conventional CRF,
 we model the conditional probability distribution of the data with the following density function:
\begin{equation}\label{eq:prob}
\begin{aligned}
\Pr(\y|\x) = \frac{1}{\mathrm{Z}(\x)} \exp \{ -E(\y, \x) \},
\end{aligned}
\end{equation}
where $E$ is the energy function and $Z(\x)$ the partition function, defined respectively as:
\begin{equation}\label{eq:feature}
E(\y, \x) = \sum_{p \in {\cal N} } (y_p - \regress_p)^2
+ \sum_{(p,q) \in {\cal S}} \half \pws_{pq} (y_p - y_q)^2,
\end{equation}
\begin{equation}\label{eq:partition}
\mathrm{Z}(\x) = \int_{\y} \exp \{ -E(\y, \x) \}\mathrm{d}\y,
\end{equation}
in which,
\begin{equation} \label{eq:def_R}
\pws_{pq} = \sum_{k=1}^K \beta_k S_{pq}^{(k)},
\end{equation}
where $\regress$ is the regressed depths parametrized by $\btheta$ (namely, $\regress$ is the abbreviation of $\regress(\btheta)$), $\bbeta=[\beta_1, \ldots, \beta_K]$ the pairwise parameter, $\S^{(k)}$ the $k$-th similarity matrix (which is symmetric) and $K$ the number of pairwise terms considered.
To guarantee $Z(\x)$ (Eq. \eqref{eq:partition}) is integrable, $\bbeta_k > 0$ are required.
We aim to jointly learn $\regress(\btheta)$ and $\bbeta$ here.

By expanding Eq. \eqref{eq:feature}, we then have:
\begin{align} \label{eq:feature_expand}
E(\y, \x) &= \sum_p y_p^2 - 2\sum_p y_p \regress_p  + \sum_p \regress_p^2 + \half \sum_{pq} \pws_{pq} y_p^2 - \sum_{pq} \pws_{pq} y_p y_q + \half \sum_{pq} \pws_{pq} y_q^2 \notag \\
	&= \y^\T \y - 2\bregress^\T \y  + \bregress^\T \bregress +  \y^\T \D \y -  \y^\T \bpws \y  \notag \\
	&= \y^\T (\I+ \D - \bpws) \y - 2\bregress^\T \y + \bregress^\T \bregress \notag \\
	&= \y^\T \A \y - 2\bregress^\T \y + \bregress^\T \bregress,
\end{align}
where
\begin{align} \label{eq:def_A}
\A = \I+ \D - \bpws.
\end{align}
Here, $\I$ is the n$\times$n identity matrix; $\D$ is a diagonal matrix with $\D_{pp} = \sum_q \pws_{pq}$.
Since $\beta_k \geq 0$ are enforced, $\A$ is ensured to be positive definite ($\A$ is symmetric, and strictly diagonally dominant with positive diagonal entries). We can then calculate the partition function according to the Gaussian integral formula as:
\begin{align} \label{eq:part_expand}
Z(\x) &=\int_\y \exp \Big\{ -E(\y, \x) \Big\}\mathrm{d}\y   \notag \\
	&=  \int_{\y} \exp \Big\{ - \y^\T \A \y + 2\bregress^\T \y - \bregress^\T \bregress \Big\}\mathrm{d}\y  \notag  \\
	&=\exp \{  - \bregress^\T \bregress \} \int_{\y} \exp \Big\{ - \y^\T \A \y + 2\bregress^\T \y \Big\}\mathrm{d}\y \notag  \\
	&= \exp \{  - \bregress^\T \bregress \}\sqrt{\frac{{(2\pi)}^n}{|2\A|}} \exp \{{ \bregress ^\T \A^{-1} \bregress}\} \notag  \\
	&= \frac{{(\pi)}^{\frac{n}{2}}}{{|\A|}^{\frac{1}{2}}} \exp \{{ \bregress ^\T \A^{-1} \bregress   - \bregress^\T \bregress }\} ,
\end{align}
where $|\A|$ denotes the determinant of the matrix $\A$, and $\A^{-1}$ the inverse of $\A$.
From Eq. \eqref{eq:prob}, \eqref{eq:feature_expand}, \eqref{eq:part_expand}, we can write the probability density function as:
\begin{align} \label{eq:prob_gaussian}
\Pr(\y|\x) &=\frac{\exp \Big\{ -E(\y, \x) \Big\}} { Z(\x)}  \\ \notag
	&= \frac{\exp \Big\{ - \y^\T \A \y + 2\bregress^\T \y - \bregress^\T \bregress \Big\}}{\frac{{(\pi)}^{\frac{n}{2}}}{{|\A|}^{\frac{1}{2}}} \exp \{{ \bregress ^\T \A^{-1} \bregress   - \bregress^\T \bregress }\}}  \\ \notag
	&=\frac{ |\A|^{\half}} {{\pi^{\frac{n}{2}}}} \exp \Big\{ - \y^\T \A \y + 2\bregress^\T \y -  \bregress ^\T \A^{-1} \bregress \Big\}. \\ \notag
\end{align}
According to Eq. \eqref{eq:prob_gaussian}, we can then rewrite the negative log-likelihood $-\log\Pr(\y|\x)$ as:
\begin{align} \label{eq:log-likelihood}
-\log\Pr(\y|\x)&= \y^\T \A \y - 2\bregress^\T \y + \bregress ^\T \A^{-1} \bregress - \half\log(|\A|) + \frac{n}{2}\log(\pi).
\end{align}

In learning, we minimizes the negative conditional log-likelihood of the training data. Adding regularization to $\btheta$, $\bbeta$, we then arrive at the final optimization:
\begin{align} \label{eq:ccrf_final}
\min_{\btheta, \bbeta \geq \zeros} &-\sum_{i=1}^N \log\Pr(\y^{(i)} | \x^{(i)}; \btheta, \bbeta) + \frac{\lambda_1}{2} \fnorm \btheta + \frac{\lambda_2}{2} \fnorm \bbeta,
\end{align}
where $\x^{(i)}$, $\y^{(i)}$ denote the $i$-th training image and the corresponding depth map; $N$ is the number of training images; $\lambda_1$ and $\lambda_2$ are weight decay parameters.

For the unary part,
here we calculate the partial derivatives of $-\log\Pr(\y|\x)$ with respect to $\theta_l$ (one element of the network parameters $\btheta$ for the unary part ). Recall that $\A = \I+ \D - \bpws$ (Eq. \eqref{eq:def_A}), $\A^{\T}=\A$, $(\A^{-1})^{\T}=\A^{-1}$, $|\A^{-1}|=\frac{1}{|\A|}$, we have:
\begin{align}  \label{eq:derive_z}
\frac{\partial \{ -\log\Pr(\y|\x)\} }{\partial \theta_l} &= \frac{\partial \{-2\bregress^\T \y + \bregress ^\T \A^{-1} \bregress\}}{\partial \theta_l}   \\ \notag
&= \frac{\partial \{-2\bregress^\T \y \} } {\partial \theta_l} + \frac{\partial \{ \bregress ^\T \A^{-1} \bregress\}}{\partial \theta_l}   \\ \notag
&= -2 \frac{\partial \{ \sum_p \regress_p y_p \} } {\partial \theta_l} + \frac{\partial \{ \sum_{pq} \regress_p \regress_q A^{-1}_{pq} \}}{\partial \theta_l}   \\ \notag
&= -2 \sum_p \Big ( y_p\frac{\partial \regress_p } {\partial \theta_l} \Big ) + \sum_{pq} \Big ( z_p \frac{\partial  \regress_q} {\partial \theta_l} + z_q \frac{\partial  \regress_p} {\partial \theta_l} \Big ) A^{-1}_{pq} \\ \notag
&= -2 \y^{\T} \frac{\partial \bregress } {\partial \theta_l} + 2\bregress^{\T} \A^{-1} \frac{\partial \bregress } {\partial \theta_l}\\ \notag
&= 2 (\A^{-1} \bregress - \y)^{\T} \frac{\partial \bregress}{\partial \theta_l} .
\end{align}


Next, for the pairwise part, we calculate the  partial derivatives of $-\log\Pr(\y|\x)$ with respect to $\beta_k$ as:
\begin{align}  \label{eq:derive_beta}
\frac{\partial \{ -\log\Pr(\y|\x)\} }{\partial \beta_k} &= \frac{\partial \{\y^\T \A \y + \bregress ^\T \A^{-1} \bregress - \half\log(|\A|)\}}{\partial \beta_k}   \notag \\
&= \frac{\partial \{ \y^\T \A \y \}}{\partial \beta_k} +  \frac{\partial \{\bregress ^\T \A^{-1} \bregress\}}{\partial \beta_k} - \half \frac{\partial \log(|\A|)}{\partial \beta_k}, \notag \\
&=\y^\T\frac{\partial  \A }{\partial \beta_k}\y - \bregress^\T \A^{-1} \frac{\partial \A }{\partial \beta_k} \A^{-1} \bregress  - \half \frac{1}{|\A|} \frac{\partial \{|\A|\} }{\partial \beta_k}, \notag \\
&=\y^\T \frac{\partial  \A }{\partial \beta_k} \y - \bregress^\T \A^{-1} \frac{\partial \A }{\partial \beta_k} \A^{-1} \bregress  - \half \trace \Big ( \A^{-1}  \frac{\partial \A} {\partial \beta_k} \Big ).
\end{align}
We here introduce matrix $\J$ to denote $\frac{\partial  \A }{\partial \beta_k}$. Each element of $\J$ is:
\begin{align}  \label{eq:def_J}
J_{pq} &= \frac{\partial A_{pq}}{\partial \beta_k}  \notag \\
&= \frac{\partial \{ D_{pq} - R_{pq} \} }{\partial \beta_k}  \notag \\
&= \frac{\partial D_{pq}  }{\partial \beta_k}  - \frac{\partial R_{pq}  }{\partial \beta_k} \notag \\
&= - \frac{\partial  \pws_{pq} }{\partial \beta_k} + \delta(p=q) \sum_q \frac{\partial  \pws_{pq} }{\partial \beta_k},
\end{align}
where $\delta(\cdot)$ is the indicator function, which equals 1 if $p=q$ is true and 0 otherwise.
From Eq. \eqref{eq:derive_beta}, Eq. \eqref{eq:def_J}, we can see that our framework is general, therefore more complicated networks for the pairwise part can be seamlessly incorporated.
Here, in our case, with the definition of $\pws_{pq}$ in Eq. \eqref{eq:def_R}, we have $\frac{\partial  \pws_{pq} }{\partial \beta_k}=S_{pq}^{(k)}$.

According to Eq. \eqref{eq:derive_beta} and the definition of  $\J$ in \eqref{eq:def_J}, we can now write the partial derivative of $-\log\Pr(\y|\x)$ with respect to $\beta_k$  as:
\begin{equation}  \label{eq:derive_beta_final}
\frac{\partial \{ -\log\Pr(\y|\x)\} }{\partial \beta_k} =  \y^{\T} \J \y - \bregress^{\T} \A^{-1} \J \A^{-1}\bregress - \half \trace \Big( \A^{-1} \J \Big).
\end{equation}

\paragraph{Depth prediction}
Predicting the depths of a new image is to solve the MAP inference.
Because of the quadratic form of $\y$ in Eq. \eqref{eq:log-likelihood}, closed form solutions exist (details refer to supplementary): 
\begin{align} \label{eq:inf_solution1}
\y^{\star}&=\argmax_{\y} \Pr(\y|\x)  \notag \\
&=\argmax_{\y} \log \Pr(\y|\x) \notag   \\
&= \argmax_{\y} -\y^\T \A \y + 2\bregress^\T \y .
\end{align}
With the definition of $\A$ in Eq.\eqref{eq:def_A}, $\A$ is symmetric.
Then by setting the partial derivative of $-\y^\T \A \y + 2\bregress^\T \y$ with respect to $\y$ to $\zeros$ ($\zeros$ is an n$\times$1 column vector with all elements being 0), we have
\begin{align} \label{eq:inf_deriv}
&\frac{\partial \{ -\y^\T \A \y + 2\bregress^\T \y \} } {\partial \y} =0  \notag  \\
\Rightarrow \;\;&-(\A + \A^{\T} ) \y +2 \bregress=0  \notag  \\
\Rightarrow \;\;&-2\A \y +2 \bregress=0  \notag  \\
\Rightarrow \;\;&\y = \A^{-1} \bregress.
\end{align}
Now we can write the solution for the MAP inference in Eq. \eqref{eq:inf_solution1} as:
\begin{align} \label{eq:inf_solution2}
\y^{\star}&= \A^{-1} \bregress
\end{align}


\section{Experiments}
\begin{figure} \center
     \includegraphics[width=0.38\textwidth, height=0.28\textwidth]{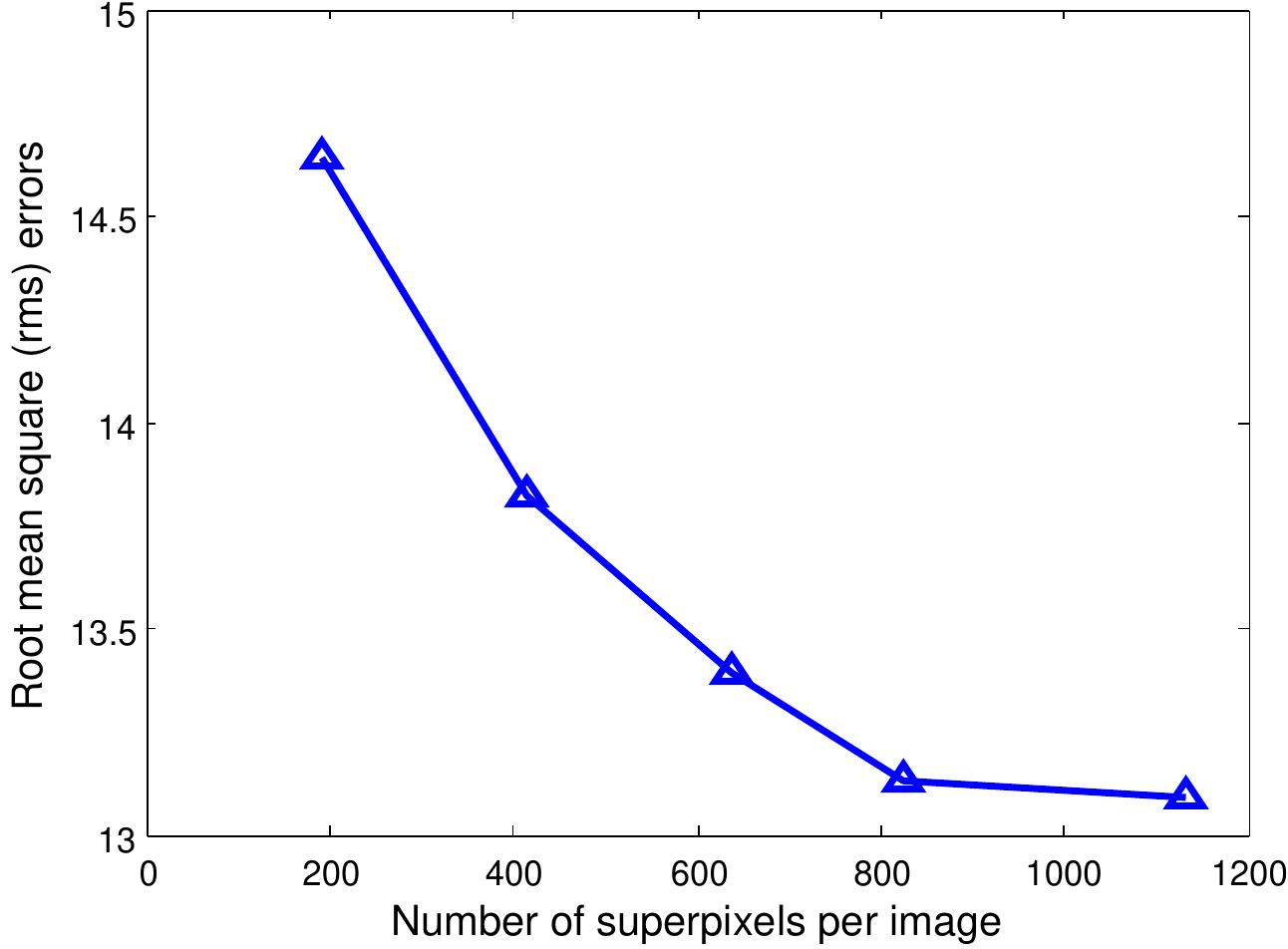}
     \includegraphics[width=0.38\textwidth, height=0.28\textwidth]{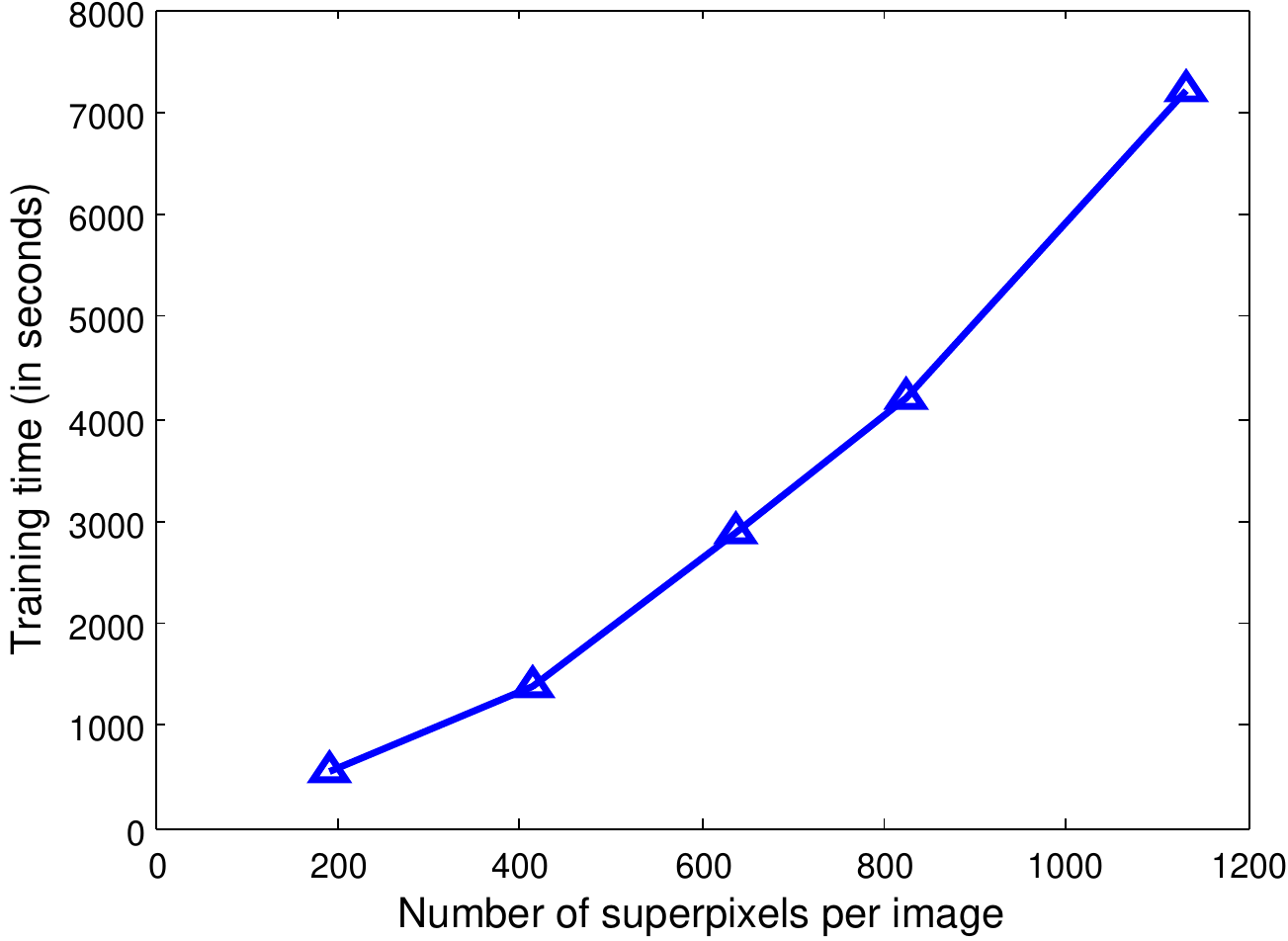}
\caption{Left: Root mean square (C2 rms)  errors  \vs  varying superpixel numbers on the Make3D dataset.
Right: Training time \vs varying superpixel numbers per image on the Make3D dataset.
Clearly, increasing the number of supperpixels per image, we can further improve the results but at the cost of more training time.}  \label{fig:rmsVSspnum}
\end{figure}

To show how the superpixel number affects the performance of our model, we add an experiment to evaluate the root mean square (rms) errors and the training time of our pre-train model on the Make3D dataset by varying the superpixel number per image.
Fig. \ref{fig:rmsVSspnum} shows the results.
As we can see, increasing the number of supperpixel per image yields further decrease in the rms error, but at the cost of more training time.
We use $\sim 700$ superpixels per image in all other experiments in this paper, therefore by increasing it, we can expect better results.

\clearpage
{\small
    \bibliographystyle{ieee-cs}
    \bibliography{CSRef}
}

\end{document}